\documentclass[letterpaper, 10 pt, conference]{ieeeconf}  % Comment this line out if you need a4paper

\IEEEoverridecommandlockouts                              % This command is only needed if 
\usepackage[pdftex]{graphicx}
\graphicspath{{figures/}}
\usepackage{hyperref}
\hypersetup{
	colorlinks=true,
	linkcolor=black,
	citecolor=black,
	filecolor=black,
	urlcolor=black,
}
\usepackage[T1]{fontenc}
\usepackage[utf8]{inputenc}
\usepackage{csquotes}
\usepackage[english]{babel}
\usepackage[export]{adjustbox}
\usepackage{caption}
\usepackage{subcaption}
\usepackage[colorinlistoftodos,backgroundcolor=orange!20,bordercolor=orange]{todonotes}
\usepackage{lipsum}
\usepackage[letterpaper,left=0.76in,right=.76in,top=0.72in,bottom=0.8in]{geometry}
\usepackage{amsmath}
\usepackage{amssymb}
\usepackage{mathtools}
\usepackage{calc}
\usepackage{pgfplots}
\usepackage{siunitx}
\usepackage{amsmath}

\DeclareMathOperator*{\argmin}{arg\,min}
\usepackage{stfloats}
\usepackage{multirow}
\usepackage{placeins}%for \FloatBarrier

\usepackage{float}
\usepackage{algorithm}
\usepackage{algorithmic}
\usepackage{booktabs}
\usepackage{svg}
\usepackage{bm}
\usepackage{amsmath} % you need amsmath as the demo includes a use of \eqref

\usepackage[style=ieee,
doi=false,
url=false,
mincitenames=1,
maxcitenames=1,
minbibnames=6,
maxbibnames=6,
backend=biber]{biblatex}  % supports bibliography with BibLaTeX
\title{\LARGE \bf TEScalib: Targetless Extrinsic Self-Calibration of LiDAR and Stereo Camera for Automated Driving Vehicles with Uncertainty Analysis}

\author{
	Haohao Hu$^{\ast,1}$, 
	Fengze Han$^{1}$,
	Frank Bieder$^{2}$,
	Jan-Hendrik Pauls$^{1}$
	and Christoph Stiller$^{1, 2}$% <-this % stops a space
\thanks{$^{\ast}$Corresponding author.}%
\thanks{$^{1}$
\href{https://www.mrt.kit.edu/}{
	Institute of Measurement and Control, 
	Karlsruhe Institute of Technology, 
	Karlsruhe, Germany. 
}
{\tt\small \{haohao.hu, jan\-hendrik.pauls, stiller\}@kit.edu, 
skyhfz@gmail.com}}%
\thanks{$^{2}$
\href{https://www.fzi.de/startseite/}{
	Mobile Perception Systems Department, 
	FZI Research Center for Information Technology, 
	Karlsruhe, Germany. 
}
{\tt\small bieder@fzi.de}}%
}

\begin{document}

\maketitle
\thispagestyle{empty}
\pagestyle{empty}
\begin{abstract}
In this paper, we present TEScalib, a novel extrinsic self-calibration approach of LiDAR and stereo camera using the geometric and photometric information of surrounding environments without any calibration targets for automated driving vehicles.
Since LiDAR and stereo camera are widely used for sensor data fusion on automated driving vehicles, their extrinsic calibration is highly important.
However, most of the LiDAR and stereo camera calibration approaches are mainly target-based and therefore time consuming.
Even the newly developed targetless approaches in last years are either inaccurate or unsuitable for driving platforms. 
To address those problems, we introduce TEScalib.
By applying a 3D mesh reconstruction-based point cloud registration, the geometric information is used to estimate the LiDAR to stereo camera extrinsic parameters accurately and robustly.
To calibrate the stereo camera, a photometric error function is builded and the LiDAR depth is involved to transform key points from one camera to another.
During driving, these two parts are processed iteratively.
Besides that, we also propose an uncertainty analysis for reflecting the reliability of the estimated extrinsic parameters.
Our TEScalib approach evaluated on the KITTI dataset achieves very promising results.
\end{abstract}
\section{INTRODUCTION}
\label{sec:INTRODUCTION}
Autonomous systems like robots and automated driving vehicles are often equipped with multi-modal sensors for environment perception.
A sensor setup composed of different sensor measuring principles make perception robust to the variety of environments.
For fusing measurements of different sensor systems, it is a key requirement to firstly obtain the extrinsic parameters between them accurately.
Most common extrinsic calibration approaches for LiDAR and stereo camera use calibration targets such as chessboards or other objects with a known shape.
The extrinsic parameters are estimated by minimizing the back-projection error of the corresponding calibration targets in the LiDAR or camera domain.
An experimental environment with enough calibration targets is usually needed, which introduces much manual work.
However, this manual work can not be easily done by customers themselves, because some professional knowledge is also needed.
Especially for custom automated driving vehicles in the future, the mounted sensors need to be re-calibrated continuously after they are sold.
\begin{figure}
    \vspace{0.2cm}
    \includegraphics[width=\columnwidth]{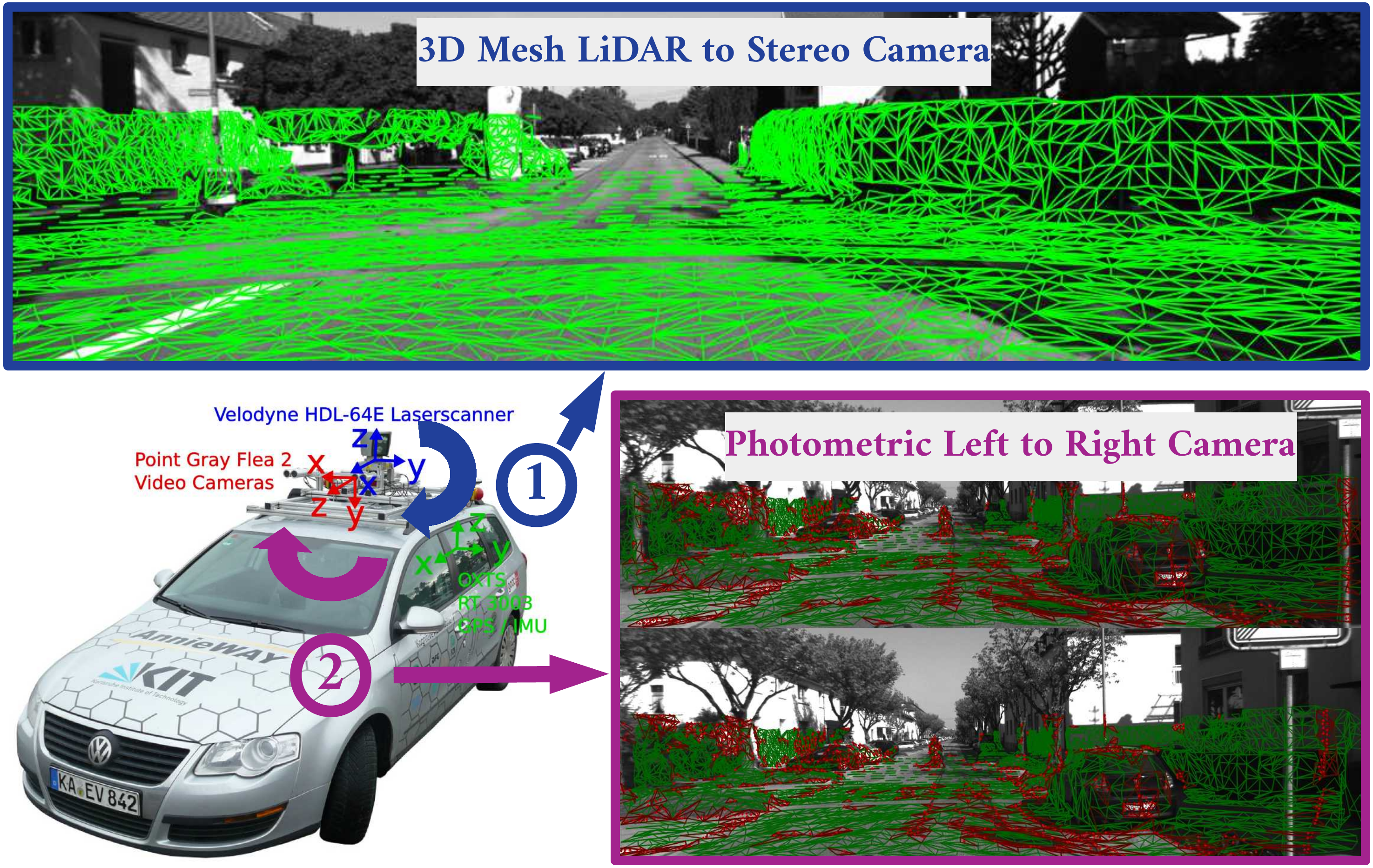}
    \caption{
        TEScalib: a targetless LiDAR stereo camera extrinsic calibration approach for automated driving platforms. 
        By applying a 3D mesh reconstruction-based point cloud registration and a photometric error function, the LiDAR and stereo camera are calibrated accurately and robustly.
    }
    \label{fig:teaser}
\end{figure}
In the work \cite{mutual}, the authors propose a targetless LiDAR and camera extrinsic calibration approach based on mutual information (MI) using LiDAR intensity measurements and camera gray values.
Cameras measure rich photometric information and LiDARs accurate geometric information of the surroundings.
Only using the inaccurate LiDAR intensity to obtain the extrinsic parameters between LiDAR and camera seems unpromising.
The work in \cite{munoz2020targetless} co-registers the camera edge pixels and LiDAR discontinuity features.
However, it faces a parallax issue since sensors can not be mounted with a same pose.
The co-registered features do not always represent a same feature in the real world.
Besides that, all targetless approaches just give us a calibration result in the end, but can not decide whether the result is reliable.
\begin{figure*}
    \vspace{0.2cm}
    \includegraphics[width=\textwidth]{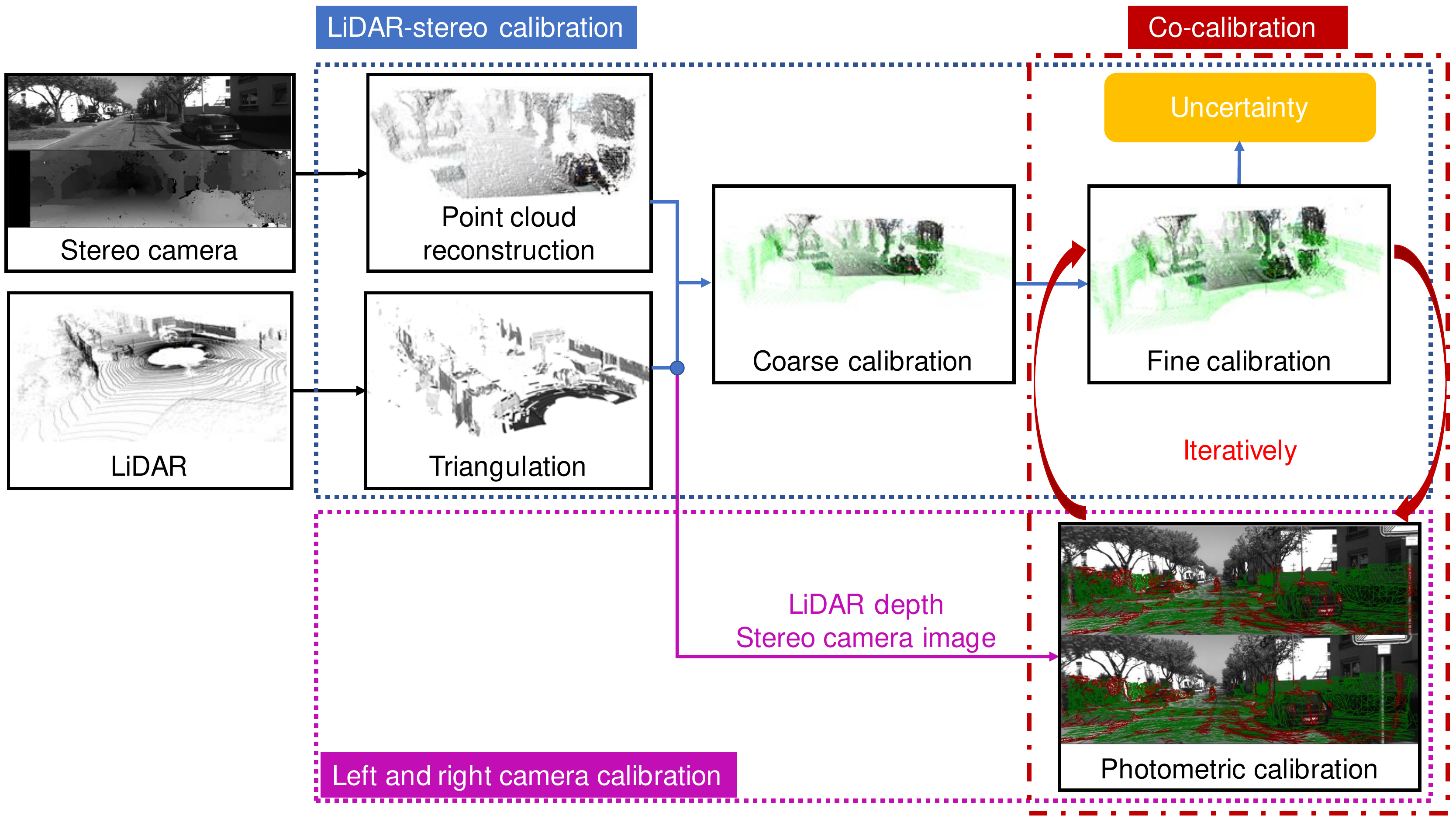}
    \caption{
        TEScalib pipeline overview.
        The LiDAR-stereo calibration (in blue) uses the environment geometric information and achieves robust and accurate results by applying a 3D mesh reconstruction-based point cloud registration method. 
        By the left and right camera calibration (in violet), camera key points and LiDAR depth are used to build a photometric error function, by minimizing which the transformation between two cameras is estimated accurately.
        These two parts run cooperatively and iteratively (in red).
        An uncertainty analysis (in orange) is done to obtain the reliability of the estimated extrinsic parameters. 
    }
    \label{fig:flowchart}
\end{figure*}
Therefore, as shown in Fig. \ref{fig:teaser} we propose TEScalib: a novel targetless extrinsic self-calibration approach of LiDAR and stereo camera using both geometric and photometric information of the surrounding environments.
Our TEScalib contains mainly four parts: LiDAR-stereo calibration, left and right camera calibration, cooperative calibration (so called co-calibration) and uncertainty analysis.
As shown in Fig. \ref{fig:flowchart}, for the LiDAR-stereo calibration, a 3D mesh reconstruction-based robust point cloud registration method is applied, which achieves a very large convergence range and a high accuracy.
A photometric error function using camera image key points and LiDAR depth is minimized to obtain the left and right camera calibration precisely.
During driving, both approaches are processed iteratively and an uncertainty analysis is involved to evaluate the reliability of the estimated parameters under different driving scenarios.
The main contributions of this work can be listed as:
\begin{itemize}
    \item LiDAR-stereo calibration using 3D mesh reconstruction achieves a large convergence range and a high accuracy.
    \item Left and right camera calibration with a photometric error function is sensitive to camera poses but not to the key point depth from the LiDAR 3D meshes, which makes an online iterative co-calibration beginning with a bad LiDAR depth initialization meaningful.
    \item An uncertainty analysis helps to evaluate the reliability of the estimated extrinsic parameters in each single dimension separately under different driving scenarios.
\end{itemize}

\section{RELATED WORK}
\label{sec:RELATED WORK}
The targetless extrinsic calibration of the LiDAR and camera sensor is the problem to estimate the rigid body transformation between the LiDAR and camera coordinate systems without any calibration targets.
The existing approaches can be mainly divided into 2 categories: the motion-based approaches and the feature-based approaches.
For the motion-based approaches like \cite{motion_based, ishikawa2018lidar}, an initialization is not required to ensure the convergence. 
Chanoh et al. \cite{spartial_temperal} use the motion estimated from a LiDAR odometry to transform camera feature points into a single coordinate system in order to build a back-projection error function, by minimizing which the extrinsic parameters between the LiDAR and camera can be obtained.
However the motion-based approaches achieve usually low accuracy because of the existing noise inside of the estimated motion. 
Besides that, in order to fix the motion degeneration problem, it requires at least 2 Degrees of Freedom (DoF) rotational motion to make all translational extrinsic parameters observable.
The feature-based approaches need an initialization and can usually achieve a higher accuracy.
In \cite{mutual, irie2016target}, the extrinsic parameters between the LiDAR and camera are estimated by maximizing the MI obtained from the sensor-measured surface intensities.
However the influence of intensities by sunlight or shadow can damage the calibration results.
The works proposed in \cite{kang2020automatic, cui2017line} estimate extrinsic parameters by aligning edges or lines detected in both LiDAR and camera domains.
In \cite{jeong2019road}, the LiDAR Simultaneous Localization and Mapping (SLAM) technique is used to map the road surface.
The mapped road surface features are then back-projected into images in order to build residuals.
The authors in \cite{shi2019extrinsic} apply a motion-based approach to estimate the extrinsic parameters coarsely and use the feature-based method to refine the result.
However finding the right feature correspondences from different sensors is a big challenge and wrong associations can strongly reduce the estimation accuracy. 
Besides that, the detected features like lines from the LiDAR and camera do not always represent the same feature in the real world due to the parallax effect.
Another big challenge of a targetless LiDAR and camera extrinsic calibration algorithm is the reliability analysis of the estimated calibration result.
Due to the lack of the constraints generated from different driving scenarios, residuals after optimization could be quite small but the elements in some specified dimensions of the calibration result could still have large errors.
The recent targetless calibration approaches lack a mechanism to make an uncertainty analysis of the estimated extrinsic parameters, which could be very helpful to evaluate the reliability of the calibration result on the one hand.
An the other hand, it is also very meaningful for an automatic selection of driving scenarios for an optimal calibration with small uncertainty in all 6 DoF. 
To address the problems, we propose TEScalib.
For the LiDAR-stereo calibration, instead of only using special features like lines or edges, we use all geometric information of driving surroundings.
No explicit association is needed and thanks to the LiDAR 3D mesh reconstruction we achieve a very large convergence and a high accuracy.
For the left and right camera calibration, the accurate LiDAR depth reduces the false key point association ratio, which makes the calibration robust and accurate.
Besides that, the uncertainty analysis evaluates the reliability of the estimated extrinsic parameters under different driving scenarios.

\section{LiDAR-stereo Calibration}
\label{sec:LiDAR-stereo Calibration}
The LiDAR-stereo calibration contains three steps: data preprocessing, coarse calibration and fine calibration.
By the data preprocessing step, the 3D environment geometry is extracted from the LiDAR and stereo camera measurements.
The goal of the coarse calibration is to estimate the extrinsic parameters with a large convergence range robustly based on a bad initialization.
Afterwards the fine calibration refines the extrinsic parameters with a very high accuracy.

\subsection{Data Preprocessing}
\label{subsec:Data Preprocessing}
To extract the 3D environment geometry from stereo camera images, a dense disparity image is calculated using the Semi-Global Block Matching (SGBM) approach presented in \cite{sgbm2005} implemented in \textit{OpenCV} \cite{bradski2008learning}.
Subsequently, we use the known camera intrinsic parameters to reconstruct the stereo camera 3D point clouds from these estimated dense disparity images.
Instead of using points directly, 3D reconstructed triangle meshes are used to represent environment geometric primitives from the LiDAR measurements, which helps us to stabilize our data association process during the coarse calibration and increase the convergence range.

\subsection{Coarse Calibration}
\label{subsec:Coarse Calibration}
\begin{figure}
    \vspace{0.2cm}
    \centering
    \begin{subfigure}{0.49\columnwidth}
        \includegraphics[width=\columnwidth]{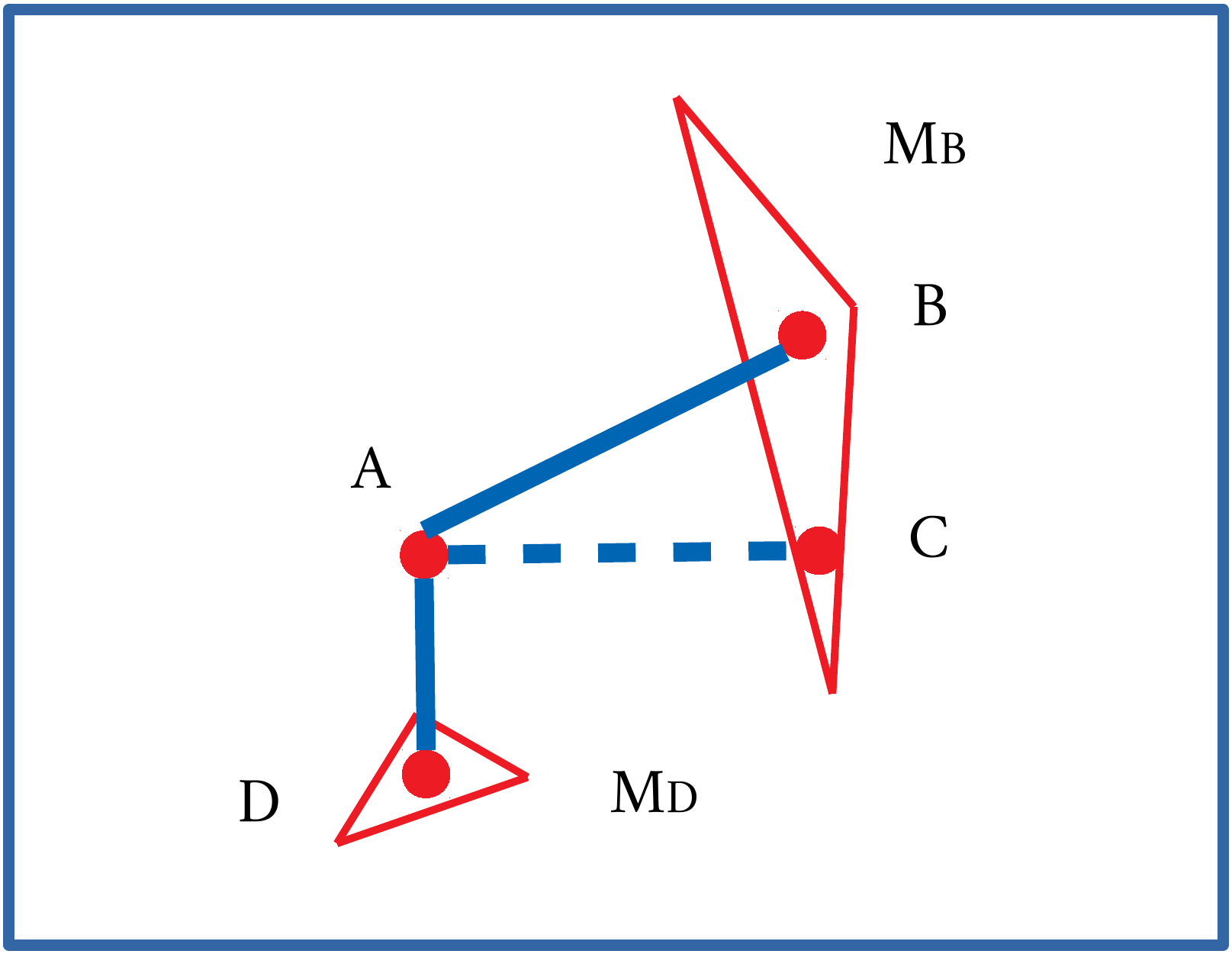}
        \subcaption{Without normals}
        \label{fig:mesh:4}
    \end{subfigure}
    \begin{subfigure}{0.49\columnwidth}
        \includegraphics[width=\columnwidth]{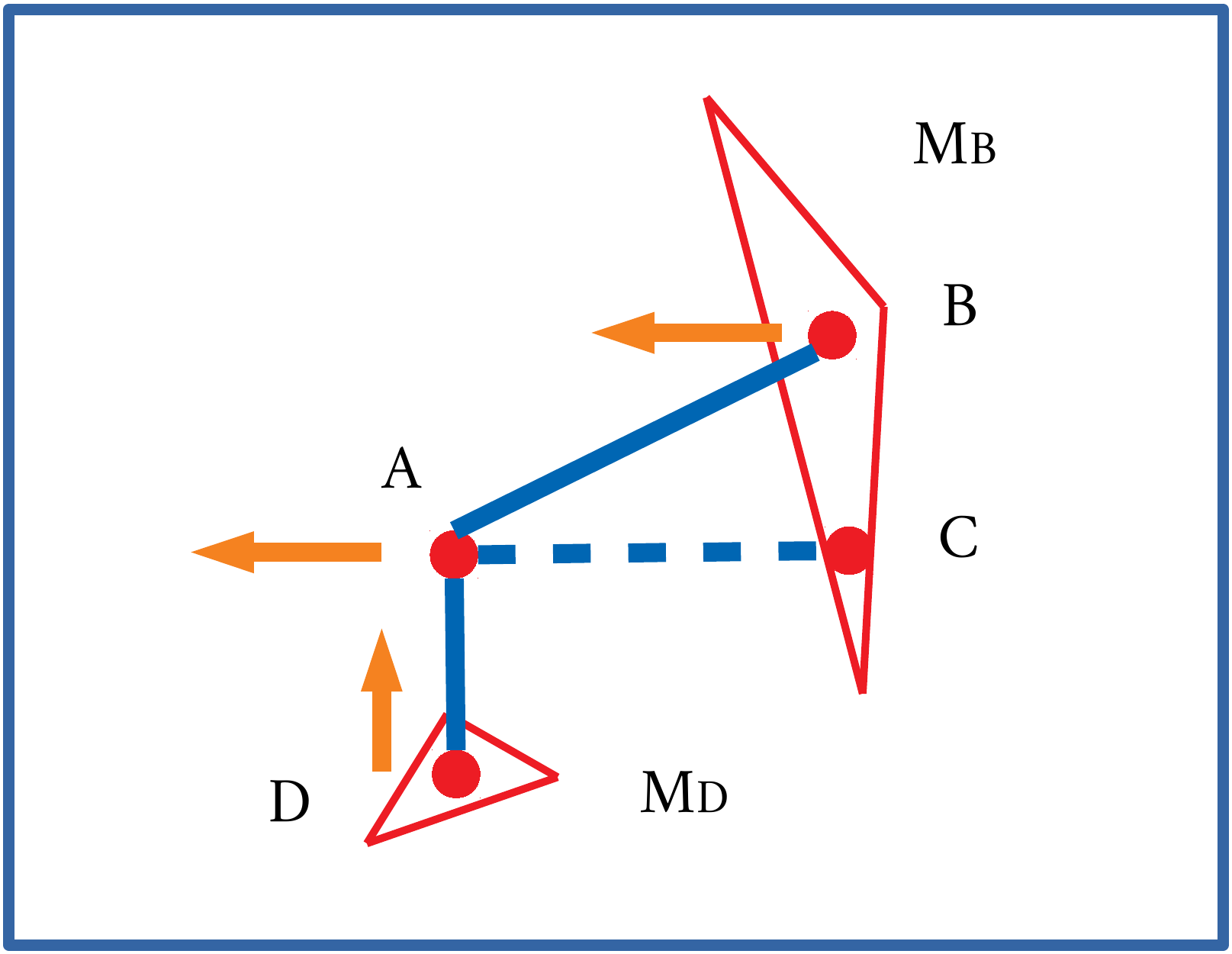}
        \subcaption{With normals}
        \label{fig:mesh:5}
    \end{subfigure}
    \caption{
    Strategies to calculate the point triangle face distance.
    }
    \label{fig:normal_association}
\end{figure}

Generally, the coarse calibration is responsible for estimating extrinsic parameters with a high convergence range and a high robustness, which is however strongly effected by the correct rate of data association.
Because of mismatches and holes in disparity images and the large initial bias of extrinsic parameters, which lead to an inaccurate association, a robust point-to-plane data association strategy is developed.
The traditional Iterative Closest Point (ICP) algorithm only uses the Euclidean distance to search the nearest neighbors and ignores the surface normal information.
Since there are many large planes in urban environments, the surface normals can provide extra information to increase the data association quality.
Therefore we add the surface normal information into our association searching process and the adapted searching criterion is defined as:
\begin{equation}
    \argmin_{^\mathrm{C}\mathbf{p}_\mathrm{i} \in \mathcal{N}_\mathrm{c}} \lVert (^\mathrm{C}\mathbf{p}_\mathrm{i} - ^\mathrm{L}\mathbf{p}_\mathrm{i}) + \omega \cdot (^\mathrm{C}\mathbf{n}_\mathrm{i} - ^\mathrm{L}\mathbf{n}_\mathrm{i})\rVert^2 \text{,}
    \label{eq:KD6}
\end{equation} 
where $\mathcal{N}_\mathrm{c}$ denotes the stereo point cloud, $^\mathrm{C}\mathbf{p}_\mathrm{i}$ denotes a stereo point and $^\mathrm{L}\mathbf{p}_\mathrm{i}$ the centroid of the corresponding LiDAR triangle face.
$^\mathrm{C}\mathbf{n}_\mathrm{i}$ is the normal vector at the stereo point and $^\mathrm{L}\mathbf{n}_\mathrm{i}$ the normal vector of the corresponding LiDAR triangle mesh.
$\omega$ is a scalar hyperparameter to balance the importance of the center point position and the normal vector.
As shown in Fig. \ref{fig:normal_association}, the surface normal information provides extra geometric information to achieve better data association results.
In Fig. \ref{fig:mesh:4}, the stereo point $A$ and centroid $D$ of the LiDAR triangle face $M_D$ have the shortest distance, if only the perpendicular distance from a point to a triangle mesh is considered.
The stereo point $A$ is then associated to the LiDAR triangle face $M_D$.
However, if the surface normal is also considered as shown in Fig. \ref{fig:mesh:5}, the stereo point $A$ will be associated to the LiDAR triangle face $M_B$ correctly.
Therefore, during the point cloud registration, we consider the surface normal information in the objective function:
\begin{equation}
    \begin{split}
        \argmin_{\mathbf{T}}\sum_{\mathbf{\mathbf{p}_\mathrm{s}^\mathrm{i},\mathbf{p}_\mathrm{m}^\mathrm{i}}} 
        \frac{\mathbf{n}_\mathrm{m}^\mathrm{i}  (^\mathrm{L}_\mathrm{C}\mathbf{T} \mathbf{p}^\mathrm{i}_\mathrm{m} - \mathbf{p}^\mathrm{i}_\mathrm{s})} {\lVert \mathbf{n}^\mathrm{i}_\mathrm{m}\rVert}
        + \mathbf{\omega} \cdot ^\mathrm{L}_\mathrm{C}\mathbf{T} \mathbf{n}^\mathrm{i}_\mathrm{m} \times \mathbf{n}^\mathrm{i}_\mathrm{s} \text{,}
    \end{split}
    \label{eq:point_plane_normal_obj}
\end{equation}
where $\mathbf{n}^\mathrm{i}_\mathrm{s}$ and $\mathbf{n}^\mathrm{i}_\mathrm{m}$ denote the normal vector of a stereo point and the normal vector of the corresponding LiDAR triangle mesh.
$^\mathrm{L}_\mathrm{C}\mathbf{T}$ denotes the LiDAR-stereo extrinsic parameters.
$\mathbf{p}^\mathrm{i}_\mathrm{s}$ is a stereo point and $\mathbf{p}^\mathrm{i}_\mathrm{m}$ the centroid point of the corresponding LiDAR triangle mesh.
Since the normal vector estimation of the ground and big walls is accurate, this objective function makes our coarse calibration accurate.

\subsection{Fine Calibration}
\label{subsec:Fine Calibration}
During the fine calibration step, a point to point objective function is developed to refine the calibration results accurately.
Because the point cloud from the stereo camera is relatively noisy, the corresponding normal vector calculation is not accurate enough, a point to point Euclidean distance is utilized to search the nearest neighbors.
Due to the high accuracy of the LiDAR depth measuring, the normal information of the LiDAR points are used to set the projection direction and the objective function is defined as:
\begin{equation}
    \begin{split}
        \argmin_\mathbf{T}\sum_{\mathbf{p}^\mathrm{i}_{\mathrm{l}},\mathbf{p}^\mathrm{i}_{\mathrm{s}}}
        {}^\mathrm{L}\mathbf{n}_\mathrm{i}^\intercal [^\mathrm{L}_\mathrm{C}\mathbf{T} ( ^\mathrm{L}\mathbf{p}_\mathrm{i} + ^\mathrm{L}\mathbf{w}_\mathrm{i})- \mathbf{F}(^\mathrm{C}d_\mathrm{i} + ^\mathrm{C}w_\mathrm{i})] \text{,}
    \end{split}
    \label{eq:point_point_normal_obj}
\end{equation}
where $\mathbf{F}(.)$ denotes the disparity to point cloud transform function, $^\mathrm{L}_\mathrm{C}\mathbf{T}$ denotes the extrinsic parameters, $^\mathrm{L}\mathbf{w}_\mathrm{i}$ and $^\mathrm{C}w_\mathrm{i}$ denote the noise of LiDAR and stereo point clouds.

\section{Left and right camera calibration}
\label{sec:Left and right camera Calibration}
An important requirement for the LiDAR-stereo calibration is the well calibrated stereo camera setup.
To ensure this, we design a stereo left and right camera extrinsic calibration approach with a camera key point-based photometric objective function using the LiDAR depth information.

\subsection{Key Point Projection Function}
\label{subsec:Key Point Projection Function}
A LiDAR triangle mesh in the 3D space can be observed from the left and right camera with the projection function:
\begin{equation}
    \begin{split}
        d_\mathrm{r} \mathbf{p}_\mathrm{r}  = \mathbf{K} \cdot ^\mathrm{r}_\mathrm{l} \mathbf{T} \left(\mathbf{K}^{-1} d_\mathrm{l} \mathbf{p}_\mathrm{l}\right) \text{,}
    \end{split}
    \label{eq:affine_transformation}
\end{equation}
where $\mathbf{p}_\mathrm{r} = \left(u_\mathrm{r}, v_\mathrm{r}, 1\right)^\intercal$ and $\mathbf{p}_\mathrm{l} = \left(u_\mathrm{l}, v_\mathrm{l}, 1\right)^\intercal$ denote the left and right camera image key point.
$d_\mathrm{r}$ and $d_\mathrm{l}$ denote the depth of the points $\mathbf{p}_\mathrm{r}$ and $\mathbf{p}_\mathrm{l}$.
$\mathbf{K}$ denotes the camera intrinsic parameters and $^\mathrm{r}_\mathrm{l}\mathbf{T}$ the stereo extrinsic parameters.
An affine relationship exists between the projected triangle in the left and right image, since $\mathbf{K}$ is not an orthonormal rather than a full range matrix.
Using the LiDAR triangle meshes, points on the triangle face can be linear interpolated with:
\begin{align}
    &\mathbf{p} = \alpha \mathbf{a} + \beta \mathbf{b} + \gamma \mathbf{c} \nonumber \\
    &\alpha + \beta + \gamma = 1 \text{,}
    \label{eq:parametric_equation}
\end{align}
where $\mathbf{a}, \mathbf{b}, \mathbf{c}$ are the triangle mesh vertices.
Because Eq.\ref{eq:parametric_equation} describes a proportional relation, we can use $\alpha, \beta ,\gamma$ to find the projection point in another camera image directly.

\subsection{Photometric Objective Function}
\label{subsec:Photometric Objective Function}
Similar as in the Stereo Direct Sparse Odometry (DSO) \cite{wang2017stereo}, the photometric objective function is denoted as: 
\begin{align}
    \mathbf{p}_\mathrm{l}&=\left[
        \begin{array}{ccc}
             \alpha & \beta & \gamma \\
        \end{array}
    \right] 
    \boxtimes \mathbf{K} \cdot ^\mathrm{C}_\mathrm{L}\mathbf{T}
    \left[
        \begin{array}{ccc}
             \mathbf{a} & \mathbf{b} & \mathbf{c} \\
             1 & 1 & 1 \\
        \end{array}
    \right] \nonumber \\
    \mathbf{p}_\mathrm{r}&=\left[
        \begin{array}{ccc}
             \alpha & \beta & \gamma \\
        \end{array}
    \right] 
    \boxtimes \mathbf{K} \cdot ^\mathrm{r}_\mathrm{l}\mathbf{T} \cdot ^\mathrm{C}_\mathrm{L}\mathbf{T}
    \left[
        \begin{array}{ccc}
             \mathbf{a} & \mathbf{b} & \mathbf{c} \\
             1 & 1 & 1 \\
        \end{array}
    \right] \nonumber \\
    E_\mathrm{lr}&=\sum_{\mathbf{p}_\mathrm{l}\in\mathcal{P}_\mathrm{l}}\sum_{\mathbf{\tilde{p}_\mathrm{l}}\in\mathcal{N}_{\mathbf{p}_\mathrm{l}}} \omega_{\mathbf{\tilde{p}_\mathrm{l}}}\lVert\mathbf{I}_\mathrm{r}\left[\mathbf{\tilde{p}_\mathrm{r}}\right] \!-\! b_\mathrm{r} \!-\! \frac{e^{a_\mathrm{r}}}{e^{a_\mathrm{l}}}\left(\mathbf{I}_\mathrm{l}\left[\mathbf{\tilde{p}_\mathrm{l}}\right] \!-\! b_\mathrm{l}\right)\rVert_\gamma \nonumber \\
    \omega_\mathbf{p}&=\frac{\mathrm{c}^2}{(\mathrm{c}^2 + \lVert \nabla \mathbf{I}\left(\mathbf{p}\right)\rVert^2_2)} \text{,} \label{eq:Photometric_object_function}
\end{align}
where $\mathbf{a}\!=\!(a_\mathrm{x}, a_\mathrm{y}, a_\mathrm{z})^\intercal$, $\mathbf{b}\!=\!(b_\mathrm{x}, b_\mathrm{y}, b_\mathrm{z})^\intercal$ and $\mathbf{c}\!=\!(c_\mathrm{x}, c_\mathrm{y}, c_\mathrm{z})^\intercal$ are the triangle face vertices,
$\alpha$, $\beta$ and $\gamma$ are the triangle face based parameterization of the point $\mathbf{p}_\mathrm{l}$,
$\boxtimes$ denotes that the last row needs to be normalized.
$\mathcal{P}_\mathrm{l}$ is the left image point set.
$\mathcal{N}_{\mathbf{p}_\mathrm{l}}$ is the used image pattern of the point $\mathbf{p}_\mathrm{l}$ to calculate photometric errors.
$\mathbf{\tilde{p}_\mathrm{r}}$ is the projection point of $\mathbf{p}_\mathrm{l}$ in the right image.
The brightness correction approach presented in \cite{wang2017stereo} is here applied and the corresponding parameters $a_\mathrm{l}$, $b_\mathrm{l}$, $a_\mathrm{r}$ and $b_\mathrm{r}$ are jointly estimated during the optimization process.
$\lVert \cdot \rVert_\gamma$ denotes the robust Huber norm and $c$ is a hyperparameter to set the weight.

\subsection{Sensitivity Analysis}
\label{subsec:Sensitivity Analysis}
To know whether the photometric objective function is suited for the left and right camera calibration using the LiDAR depth in the camera frame, which could be noisy because of a bad LiDAR-stereo calibration initialization, we need to analyze the sensitivity of the objective function to the camera pose and LiDAR depth.
We assume that the depth of the same key point in the left and right camera is the same: $d = d_\mathrm{l} = d_\mathrm{r}$.
The sensitivity depends on Eq. \ref{eq:Photometric_object_function} to the camera pose $^\mathrm{r}_\mathrm{l}\mathbf{T}$ and to the LiDAR depth $d$ are denoted as:
\begin{align}
    \mathbf{J}_{^\mathrm{r}_\mathrm{l}\mathbf{T}} &= \frac{1}{d_\mathrm{r}}\mathbf{K} \left[^\mathrm{r}_\mathrm{l}\mathbf{T}\left(\mathbf{K}^{-1}d_\mathrm{l}\mathbf{p}_\mathrm{l}\right), \mathbf{I}\right] \in \mathbb{R}^{3 \times 6} \nonumber \\ 
    \mathbf{J}_{d_\mathrm{r}} &= \frac{1}{d_\mathrm{r}^2} \mathbf{K} \ ^\mathrm{r}_\mathrm{l}\mathbf{R} \mathbf{p}_\mathrm{l} \in \mathbb{R}^{3 \times 1} \text{.} \label{eq:stereo_sensitivity}
\end{align}
By applying camera intrinsics, the rotation and translation Jacobian of the camera pose and key point depth are then:
\begin{align}
    \mathbf{J}_{^\mathrm{r}_\mathrm{l}\mathbf{R}} &= \left[
    \begin{array}{ccc}
         \!c_\mathrm{x} v\!  &  \!f_\mathrm{x} \left(\frac{t_\mathrm{x}}{d} \!-\! u\right) \!+\! c_\mathrm{x}\!  &  \!f_\mathrm{x} v\! \\
         \!f_\mathrm{y} \left(\frac{t_\mathrm{x}}{d} \!-\! u\right) \!-\! c_\mathrm{y}\!  &  \!c_\mathrm{y}\!  &  \!f_\mathrm{y}\! \\
         \!v\!  &  \!1\!  &  \!0\!  \\
    \end{array}
    \right] \nonumber \\
    \mathbf{J}_{^\mathrm{r}_\mathrm{l}\mathbf{T}} &= \left[
    \begin{array}{ccc}
         \frac{f_\mathrm{x}}{d}  &  0  &  \frac{c_\mathrm{x}}{d} \\
         0  &  \frac{f_\mathrm{y}}{d}  &  \frac{c_\mathrm{y}}{d} \\
         0  &  0  &  \frac{1}{d} \\
    \end{array}
    \right] \nonumber \\
    \mathbf{J}_{d_\mathrm{r}} &= \frac{1}{d_\mathrm{r}^2}\left[
    \begin{array}{c}
         f_\mathrm{x} u + c_\mathrm{x} \\
         f_\mathrm{y} v + c_\mathrm{y} \\
         1 \\
    \end{array}
    \right] \text{.}
    \label{eq:stereo_sensitivity_jacobian}
\end{align}
As shown in Eq. \ref{eq:stereo_sensitivity_jacobian}, the photometric objective function is sensitive to the camera pose but not to the LiDAR depth.
Therefore the photometric objective function can tolerant some LiDAR stereo calibration noise and still realize an accurate left and right camera calibration.

\section{Co-calibration}
\label{sec:Co-calibration}
The LiDAR-stereo calibration is robust and accurate, when the stereo camera setup is well calibrated.
The left and right camera calibration is accurate and can tolerant some LiDAR stereo camera extrinsic parameter error.
During driving, these two approaches are processed cooperatively and iteratively, which is called as co-calibration.
We stop this process until both of the extrinsic parameter estimations are stable.

\section{Uncertainty Analysis}
\label{sec:Uncertainty Analysis}
Most of LiDAR and camera calibration approaches can not decide automatically whether the calibration result is reliable when the residuals are small after optimization.
We propose a novel uncertainty analysis to evaluate the reliability of the estimated extrinsic parameters under different scenarios.

\subsection{Measurement Noise Model}
\label{subsec:Measurement Noise Model}
Measurement noise leads to the uncertainty of the extrinsic parameter estimation.
To analyze the estimation uncertainty, sensor noise models are needed.
The stereo point cloud is obtained from the disparity image.
Let $^\mathrm{C}\mathrm{w}_\mathrm{i} \in \mathcal{N}(0, ^\mathrm{C}\Sigma_\mathrm{i})$ be the disparity noise of each pixel.
The covariance matrix is $^\mathrm{C}\Sigma_\mathrm{i} = \sigma_\mathrm{i}^2  \mathbf{I}_{1 \times 1}$.
The disparity of each pixel can be described as $^\mathrm{C}d^\mathrm{gt}_\mathrm{i} = ^\mathrm{C}d_\mathrm{i} + ^\mathrm{C}\mathrm{w}_\mathrm{i}$.
Similar as \cite{yuan2021pixel}, let $^\mathrm{L}\mathbf{w}_\mathrm{i}$ be the measurement noise of the LiDAR point $^\mathrm{L}\mathbf{p}_\mathrm{i}$.
Based on the measuring principle of LiDAR sensors, this noise can be divided into the beam direction noise and the distance noise assuming that the beam direction has a perturbation in the tangent plane.
Let $^\mathrm{l}\mathbf{w}_\mathrm{i} \in \mathcal{N}(\mathbf{0}, ^\mathrm{l}\mathbf{\Sigma}_\mathrm{i})$ be the beam direction noise.
The covariance matrix is $^\mathrm{l}\mathbf{\Sigma}_\mathrm{i} = \sigma_\mathrm{i}^2  \mathbf{I}_{2 \times 2}$.
The beam direction $\bm{\omega}_{\mathrm{i}}^\mathrm{gt} = e^{\lfloor \mathrm{N}(\bm{\omega}_\mathrm{i}) ^\mathrm{l}\mathbf{w}_\mathrm{i} \times \rfloor} \bm{\omega}_{\mathrm{i}}$, where $\lfloor \; \times \rfloor$ denotes the skew-symmetric matrix mapping, $\mathrm{N}(\bm{\omega}_\mathrm{i}) = [\mathbf{N}_1, \mathbf{N}_2] \in \mathbb{R}^{3 \times 2}$ is the orthonormal basis of the tangent space at $\bm{\omega}_\mathrm{i}$.
Let $^\mathrm{d}\mathrm{w}_\mathrm{i} \in \mathcal{N}(0, ^\mathrm{d}\Sigma_\mathrm{i})$ be the distance noise, where $^\mathrm{d}\Sigma_\mathrm{i} \in \mathbb{R}^{1 \times 1}$.
The distance is described as $^\mathrm{l}d^\mathrm{gt}_\mathrm{i} = ^\mathrm{l}d_\mathrm{i} + ^\mathrm{d}\mathrm{w}_\mathrm{i}$.
Combining these two noise sources, the ground-truth LiDAR point $^\mathrm{L}\mathbf{p}^\mathrm{gt}_\mathrm{i}$ combined with its measurement $^\mathrm{L}\mathbf{p}_\mathrm{i}$ can be described as:
\begin{align}
    ^\mathrm{L}\mathbf{p}^\mathrm{i}_\mathrm{gt} &= ^\mathrm{l}d^\mathrm{gt}_\mathrm{i} \cdot \bm{\omega}^{\mathrm{gt}}_\mathrm{i} \nonumber \\
    & \approx \underbrace{^\mathrm{l}d_\mathrm{i} \cdot \bm{\omega}_\mathrm{i}}_{^\mathrm{L}\mathbf{p}_\mathrm{i}} + \underbrace{^\mathrm{d}\mathrm{w}_\mathrm{i} \cdot \bm{\omega}_{\mathrm{i}} - ^\mathrm{l}d_\mathrm{i} \lfloor \bm{\omega}_\mathrm{i} \times \rfloor \mathbf{N}(\bm{\omega}_i) ^\mathrm{l}\mathbf{w}_\mathrm{i}}_{^\mathrm{L}\mathbf{w}_\mathrm{i}} \text{.}
    \label{eq:lidar_point_gt}
\end{align}
Therefore, the covariance matrix of the LiDAR point measurement noise $^\mathrm{L}\mathbf{w}_\mathrm{i} \in \mathcal{N}(\mathbf{0}, ^\mathrm{L}\mathbf{\Sigma}_\mathrm{i})$ can be described as:
\begin{align}
    ^\mathrm{L}\mathbf{\Sigma}_\mathrm{i} &= \mathbf{J}_\mathrm{i} \left[
    \begin{array}{cc}
        ^\mathrm{d}\mathrm{\Sigma}_\mathrm{i} & \mathbf{0}_{1 \times 2} \\
        \mathbf{0}_{2 \times 1} & ^\mathrm{l}\mathbf{\Sigma}_\mathrm{i} \\
    \end{array}
    \right] \mathbf{J}_\mathrm{i}^\intercal \nonumber \\
    \mathbf{J}_\mathrm{i} &= [\bm{\omega}_\mathrm{i},\; - ^\mathrm{l}d_\mathrm{i} \lfloor \bm{\omega}_\mathrm{i}\times \rfloor \mathbf{N}(\bm{\omega}_\mathrm{i})] \in \mathbb{R}^{3 \times 3} \text{.}
    \label{eq:lidar_noise}
\end{align}

\subsection{Uncertainty Analysis under Different Driving Scenarios}
\label{subsec:Uncertainty Analysis under Different Driving Scenarios}
Based on the constraints generated from different driving scenarios, the estimated extrinsic parameters can have high accuracy in some dimensions and low accuracy in other dimensions.
Therefore, we introduce a novel uncertainty analysis approach to evaluate the reliability of the LiDAR-stereo calibration result.
The corresponding objective function Eq.\ref{eq:point_point_normal_obj} can be approximated as:
\begin{align}
    0 &= ^\mathrm{L}\mathbf{n}_\mathrm{i}^\intercal [^\mathrm{L}_\mathrm{C}\mathbf{T} (^\mathrm{L}\mathbf{p}_\mathrm{i} + ^\mathrm{L}\mathbf{w}_\mathrm{i})- \mathbf{F}(^\mathrm{C}d_\mathrm{i} + ^\mathrm{C}w_\mathrm{i})] \nonumber \\
    &\approx \mathbf{res}_\mathrm{i} + \mathbf{J}_{\mathbf{T}_\mathrm{i}} \bm{\delta} ^\mathrm{L}_\mathrm{C}\mathbf{T} + \mathbf{J}_{\mathbf{w}_\mathrm{i}} \delta \mathbf{w}_\mathrm{i} \text{,}
    \label{eq:residual_approximation}
\end{align}
where
\begin{align}
    \mathbf{res}_\mathrm{i} &= ^\mathrm{L}\mathbf{n}_\mathrm{i}^\intercal [^\mathrm{L}_\mathrm{C}\mathbf{T} (^\mathrm{L}\mathbf{p}_\mathrm{i}) - \mathbf{F}(^\mathrm{C}d_\mathrm{i})] \nonumber \\
    ^\mathrm{L}_\mathrm{C}\mathbf{T} &=   
    \left[
        \begin{array}{cc}
             \mathbf{R}  &  \mathbf{t}  \\
             \mathbf{0}_{3 \times 1}  &  \mathrm{0}_{1 \times 1}  \\
        \end{array}
    \right] \nonumber \\
    \mathbf{J}_{\mathbf{T}_\mathrm{i}}  &=  ^\mathrm{L}\mathbf{n}_\mathrm{i}^\intercal \left[ - \lfloor\mathbf{R} ^\mathrm{L}\mathbf{p}_\mathrm{i} \rfloor ,\; \mathbf{I} \right] \nonumber \\
    \mathbf{J}_{\mathbf{w}_\mathrm{i}}  &=  ^\mathrm{L}\mathbf{n}_\mathrm{i}^\intercal \left[ \mathbf{R}, -\frac{\partial \mathbf{F}}{\partial d_\mathrm{i}} \right] \nonumber \\
    \mathbf{w}_\mathrm{i}  &=  \left[ 
        \begin{array}{c}
             ^\mathrm{L}\mathbf{w}_\mathrm{i}  \\
             ^\mathrm{C}\mathrm{w}_\mathrm{i}  \\
        \end{array}
    \right]  \in  \mathbb{R}^{4  \times  1} \nonumber \\
    \bm{\Sigma}_\mathrm{i}  &=  \left[
        \begin{array}{cc}
             ^\mathrm{L}\mathbf{\Sigma}_\mathrm{i}  &  \mathbf{0}  \\
             \mathbf{0}  &  ^\mathrm{C}\mathrm{\Sigma}_\mathrm{i}  \\
        \end{array}
    \right]  \in  \mathbb{R}^{4  \times  4} \text{.}
    \label{eq:residual_explain}
\end{align}
Assuming no bias in the left and right camera calibration, the uncertainty of the LiDAR-stereo calibration only depends on the noise of LiDAR points and pixel disparities.
Based on the forward and backward propagation, the LiDAR-stereo calibration uncertainty can be defined as:
\begin{equation}
    \begin{split}
        ^\mathrm{L}_\mathrm{C}\mathbf{\Sigma}_\mathbf{T} = \left[ \sum_{^\mathrm{L}\mathbf{p}_\mathrm{i}, ^\mathrm{C}d_\mathrm{i}} \mathbf{J}_{\mathbf{T}_\mathrm{i}}^\intercal  \left( \mathbf{J}_{\mathbf{w}_\mathrm{i}} \mathbf{\Sigma}_\mathrm{i} \mathbf{J}_{\mathbf{w}_\mathrm{i}}^\intercal \right)^{-1} \mathbf{J}_{\mathbf{T}_\mathrm{i}} \right]^{-1} \text{.}
    \end{split}
    \label{eq:residual_uncertainty}
\end{equation}
\section{EXPERIMENTAL EVALUATION}
\label{sec:EXPERIMENTAL EVALUATION}
\subsection{Dataset and Parameters}
\label{sec:Dataset and Parameters}
For the experimental evaluation, we use the urban sequence 00 from the KITTI dataset \cite{geiger2013vision}.
For comparison, the ground-truth extrinsic parameters between the LiDAR and stereo camera setup as well as between the left and right stereo camera provided by the KITTI dataset are used.
We also compare our LiDAR-stereo calibration results based on the provided left and right camera ground-truth calibration against other state-of-art targetless approaches, which are also evaluated on the KITTI dataset.
Besides that, the co-calibration, which combines these two approaches, is also evaluated.
We evaluate our uncertainty analysis of the estimated extrinsic parameters under some typical driving scenarios and explain the meaning of the estimated uncertainty.
The evaluation shows that our approaches are accurate and robust by applying urban driving scenarios.

\subsection{Results and Analysis}
\label{sec:Results and Analysis}
\subsubsection{LiDAR-stereo calibration}
\label{subsec:LiDAR-stereo calibration}
\begin{figure}
    \vspace{0.2cm}
    \begin{subfigure}{\columnwidth}
        \includegraphics[width=\columnwidth]{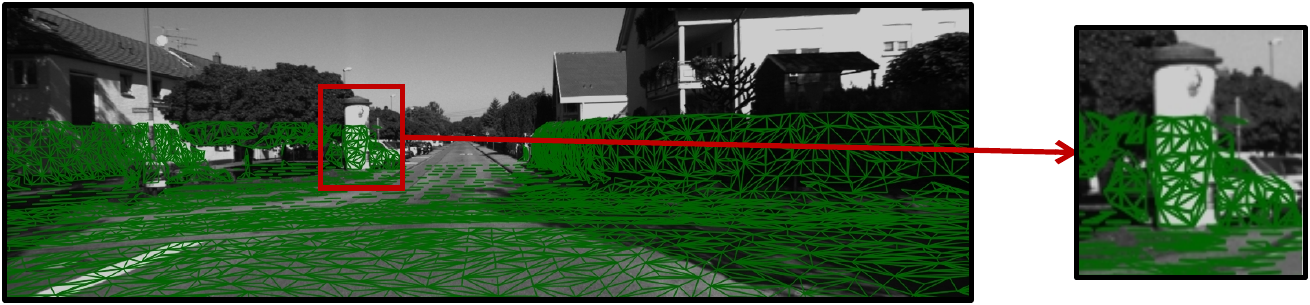}
        \subcaption{The reprojection error in the image domain}
        \label{fig:fine_calib}
    \end{subfigure} 
    \begin{subfigure}{\columnwidth}
        \includegraphics[width=\columnwidth]{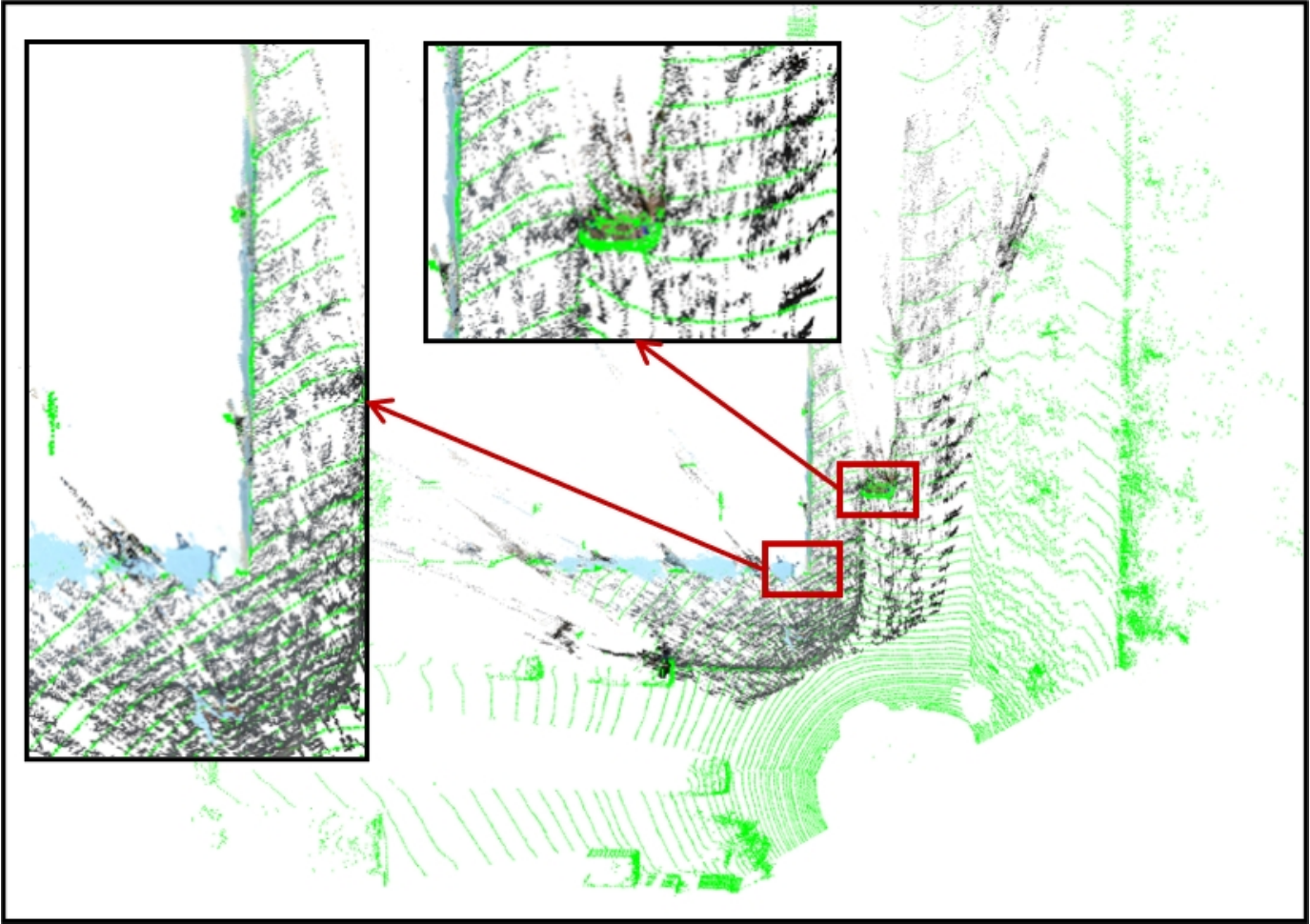}
        \subcaption{The distance error in the 3D space}
        \label{fig:3D_calib}
    \end{subfigure}
    \caption{
        An LiDAR-stereo calibration error example tested on the KITTI dataset in the image domain and the 3D space.
        The green points and meshes denote the LiDAR points and triangle faces.
        The black points denote the stereo points.
    }
    \label{fig:reprojection_result}
\end{figure}

To use the approach presented in Sec. \ref{subsec:Coarse Calibration} and Sec. \ref{subsec:Fine Calibration} to extrinsic calibrate the LiDAR and stereo camera setup, a batch optimization with a batch size of 20 is applied.
This procedure is processed as a sliding window approach over time of a driving sequence.
To use measurements with meaningful information, we select measurements with a time interval of $0.5$ s.
The rotation initialization error is gaussian randomly simulated with a mean error of $3.0^\circ$ and the translation initialization error with a mean error of $0.3$ m.
The hyperparameter to balance the point coordinates and normal vector $\omega$ in Eq.\ref{eq:KD6} is chosen as $0.5$.
The calibration results based on an the KITTI sequence 00 are plotted in Fig. \ref{fig:calibration_result} and the comparison of our calibration results on the KITTI dataset against other state-of-the-art approaches are shown in TABLE. \ref{table:calibration_result}.
As shown in Fig. \ref{fig:calibration_result}, the coarse calibration is robust against bad initializations.
After the coarse calibration, the rotation error of the calibration result is smaller than $0.6^{\circ}$ and the translation error smaller than $0.2$ m.
After the fine calibration, the rotation error of the result is smaller than $0.1^\circ$ and the translation error about $0.01$ m.
The estimated extrinsic parameters between the LiDAR and stereo camera have a lower covariance and our approach has a high consistency.
Besides that, as shown in TABLE. \ref{table:calibration_result}, our two-stage LiDAR-stereo calibration approach outperforms all of other state-of-the-art targetless extrinsic calibration approaches.
A calibration result example tested on the KITTI dataset is illuminated in Fig. \ref{fig:reprojection_result}.
Fig. \ref{fig:fine_calib} shows the reprojection error on the image plane and both of the pillar edges are matched accurately.
The calibration results are also shown in Fig.\ref{fig:3D_calib} in the 3D space.
The wall and vehicles in the LiDAR and stereo point clouds are aligned also accurately.
This example proves that our approach achieves high-accurate results.

\subsubsection{Left and right camera calibration}
\begin{figure*}
    \vspace{0.2cm}
    \includegraphics[width=0.98\textwidth]{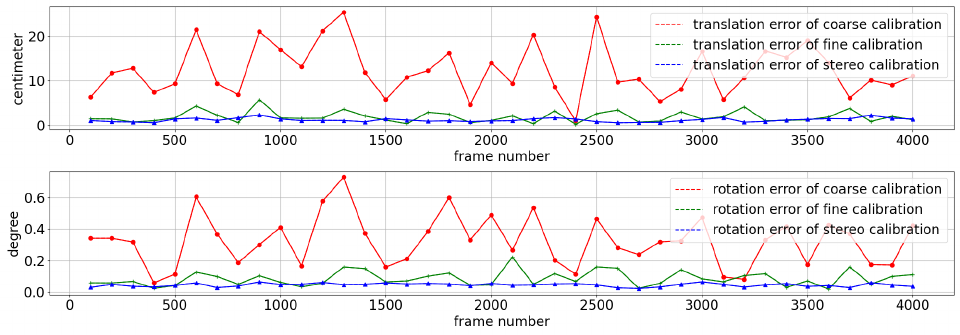}
    \caption{
        The LiDAR-stereo as well as the left and right camera calibration results evaluated on the sequence 00 of the KITTI dataset.
        The red line shows the LiDAR-stereo coarse calibration result errors (gaussian randomly initialized with a rotation mean error of $3.0^\circ$ and a translation mean error of $0.3$ m).
        The blue line shows the left and right camera calibration result errors (gaussian randomly initialized with a rotation mean error of $1.0^\circ$ and a translation mean error of $0.1$ m).
    }
    \label{fig:calibration_result}
\end{figure*}

In this section, we evaluate the left and right camera calibration presented in Sec. \ref{subsec:Key Point Projection Function} and Sec. \ref{subsec:Photometric Objective Function}.
We assume that the transformation between the LiDAR and stereo camera setup is known.
As same as the LiDAR-stereo calibration, we apply a batch optimization with a batch size of $20$ and the measurements are sampled with a time distance of $0.5$ s.
The initial rotation and translation error are gaussian randomly simulated with a mean errors of $1.0^{\circ}$ and $0.1$ m.
The calibration results evaluated on the KITTI sequence 00 are shown in Fig. \ref{fig:calibration_result} in blue.
Our approach achieves a high-level accurate performance.
The most rotation error is smaller than $0.1^\circ$ and the translation error smaller than $0.01$ m.
\begin{table}
    \centering
    \vspace{0.2cm}
    \caption{
        Comparing our LiDAR-stereo calibration with other state-of-the-art approaches like HEC\cite{HandEyeCalib}, STC\cite{Spatiotemporal}, SOIC \cite{xiao2017accurate}, MBC \cite{wang2020soic} and ECM \cite{taylor2016motion} tested on the KITTI dataset.
        Our approach outperforms all of these approaches.
    }
    \begin{tabular}{ccccccc}
    \hline
           & Rx$(^\circ)$ & Ry$(^\circ)$ & Rz$(^\circ)$ & Tx(cm)   & Ty(cm)   & Tz(cm)   \\ \hline
    Ours   & -0.002       & -0.012       & 0.037        & 0.664    & -0.513   & 0.331    \\
    HEC    & 0.800        & -5.360       & 1.600        & 1.460    & 13.310   & 56.970   \\
    STC    &-0.390        & 0.010        & 0.060        & -2.150   & -31.710  & 9.800    \\
    SOIC   & 0.070        & -0.170       & -0.230       & 6.100    & -8.600   & 9.000    \\
    MBC    & 0.438        & 0.415        & 0.154        & 6.300    & 7.500    & 2.000    \\
    ECM    & 0.353        & 0.283        & 0.313        & 2.300    & 3.800    & 2.300    \\
    ECE    & 0.042        & 0.030        & -0.040       & -0.520   & 0.700    & 1.310    \\ \hline
    \label{table:calibration_result}
    \end{tabular}
\end{table}
\begin{figure}
    \includegraphics[width=\columnwidth]{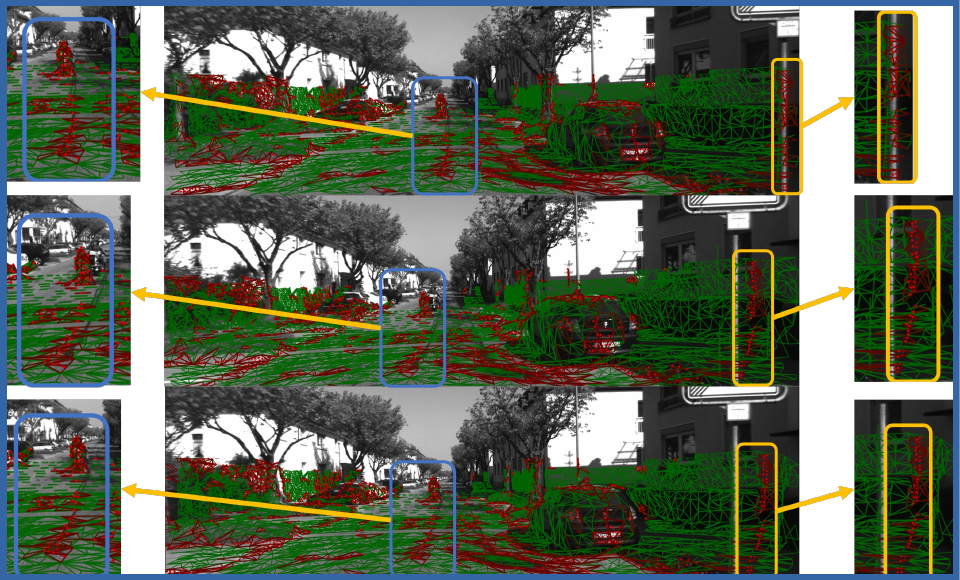}
    \caption{
        The left and right camera calibration using a photometric error function.
        The upper image comes from the left camera.
        The middle and lower image come from the right camera before and after calibration.
        The key points in the blue rectangle are aligned accurately after the optimization.
    }
    \label{fig:photo_evaluation}
\end{figure}

\subsubsection{Co-calibration}
\label{subsec:Co-calibration}
\begin{figure*}
    \vspace{0.2cm}
    \includegraphics[width=0.98\textwidth]{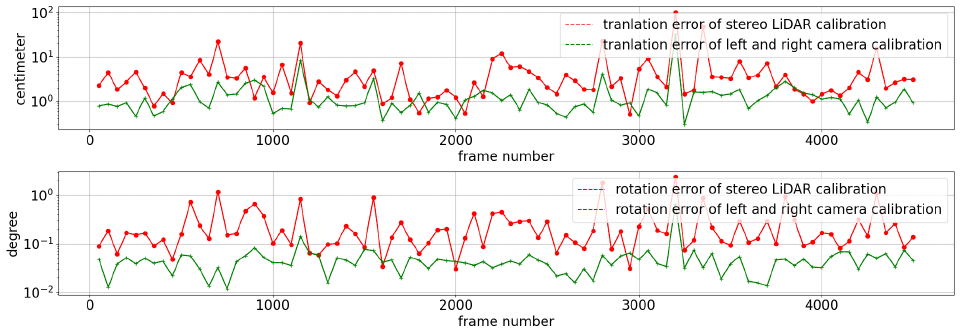}
    \centering
    \caption{
        The co-calibration results evaluated on the sequence 00 of the KITTI dataset.
        The red line denotes the LiDAR-stereo calibration result errors and the blue line the left and right camera calibration result errors.
        The initialization of the co-calibration is gaussian randomly simulated with a rotation mean error of $1.0^\circ$ and a translation mean error of $10.0$ cm).
    }
    \label{fig:co_calibration}
\end{figure*}

In this section, we evaluate the co-calibration process, which estimates the LiDAR-stereo extrinsic parameters and the left and right camera extrinsic parameters iteratively.
Firstly, we start with the left and right camera calibration using the LiDAR depth information, because as explained in Sec. \ref{subsec:Sensitivity Analysis} this approach is not sensitive to the key point depth noise.
And then we run the LiDAR-stereo calibration with the estimated stereo camera calibration.
Afterwards, the estimated LiDAR-stereo extrinsic parameters are fed into the left and right camera calibration process.
This operation is repeated until both of these two calibration processes are converged.
The batch optimization has a batch size of $10$ and a sampling interval of $0.5$ s.
The initial rotation and translation error is gaussian randomly simulated with a mean errors of $1.0^{\circ}$ and $0.1$ m.
The calibration results evaluated on the KITTI sequence 00 are shown in Fig. \ref{fig:co_calibration}.
For the left and right camera calibration, most of the rotation error is smaller than $0.1^\circ$ and the translation error smaller than $0.02$ m.
For the LiDAR-stereo camera calibration, most of the rotation error is smaller than $0.3^\circ$ and the translation error smaller than $0.05$ m.

\subsection{Uncertainty under Different Driving Scenarios}
\label{subsec:Uncertainty under Different Driving Scenarios Eva}
\begin{figure}
    \begin{subfigure}{\columnwidth}
        \includegraphics[width=\columnwidth]{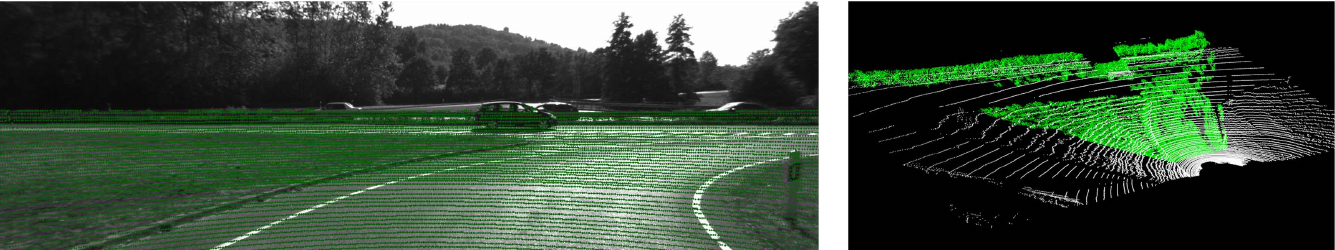}
        \subcaption{Highway scenario}
        \label{fig:highway}
    \end{subfigure}
     \begin{subfigure}{\columnwidth}
        \includegraphics[width=\columnwidth]{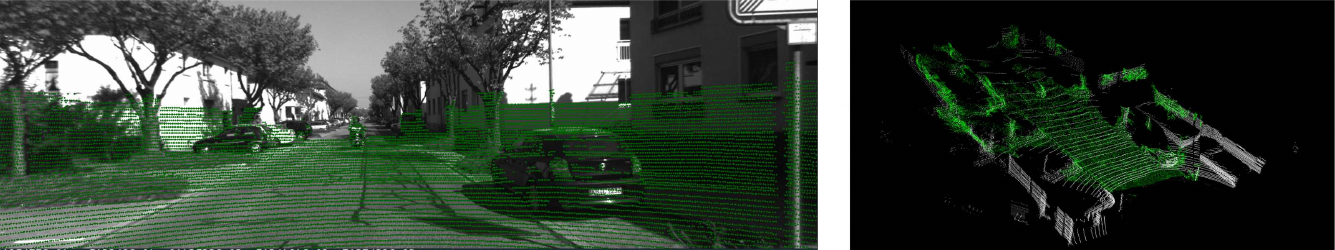}
        \subcaption{Urban scenario}
        \label{fig:urban}
    \end{subfigure}
    \centering
    \caption{
        Uncertainty analysis under two scenarios.
        The stereo points are green colored and the LiDAR points white colored.
    }
    \label{fig:uncertainty_scenario}
\end{figure}

The targetless calibration approaches can provide extrinsic parameters automatically without any calibration targets.
However, most of the targetless approaches have limitations on suitable driving scenarios.
Using the uncertainty analysis, the estimation reliability of the extrinsic parameters under different scenarios can be obtained after optimization.
We select 2 typical scenarios for the uncertainty analysis evaluation of our LiDAR-stereo calibration approach.
The Fig. \ref{fig:highway} and Fig. \ref{fig:urban} denote a highway and an urban environment from the KITTI dataset.
The calibration errors from the highway and urban environment are shown in Eq.\ref{eq:stereo_uncertainty_highway} and Eq. \ref{eq:stereo_uncertainty_urban}.
In the highway scenario, the translation error along the x- and z-axis as well as the rotation error along the y-axis are larger than the error in other dimensions.
It is actually easy to understand.
In this scenario, we try to align 2 planes.
If these two planes have a big translation difference along the x- and z-axis and a big rotation difference along the y-axis, the residuals are still small.
It means that the reliability of the calibrated translation result along the x- and z-axis and the rotation result along the y-axis is poor.
\begin{align}
    \mathbf{E}_\mathbf{R} &= \left(
    \begin{array}{ccc}
         0.2116  &  1.8645  &  0.0398  \\
    \end{array}
    \right) (^\circ) \nonumber \\
    \mathbf{E}_\mathbf{T} &= \left(
    \begin{array}{ccc}
         -1.3658 	&  0.0218  &  -2.2419  \\
    \end{array}
    \right) (\text{m})
    \label{eq:stereo_uncertainty_highway}
\end{align}
In the urban scenario, the error in each dimension is small, because the scenario provides enough constraints in each direction.
Our algorithm achieves accurate and reliable calibration result by applying constraints from this scenario.
\begin{align}
    \mathbf{E}_\mathbf{R} &= \left(
    \begin{array}{ccc}
         0.036  &  -0.042  &  0.042  \\
    \end{array}
    \right) (^\circ) \nonumber \\
    \mathbf{E}_\mathbf{T} &= \left(
    \begin{array}{ccc}
         0.004  &  0.015  &  0.006  \\
    \end{array}
    \right) (\text{m})
    \label{eq:stereo_uncertainty_urban}
\end{align}
The uncertainty ratio (the uncertainty under the highway scenario element wise divided by the uncertainty under the urban scenario) shown in Eq. \ref{eq:uncertainty_ratio} reflects the calibration reliability under these two driving scenarios.
\begin{align}
    \bm{\Sigma}_\mathbf{R} &= \left(
    \begin{array}{ccc}
         0.705 	   &  -120.293  &  -0.308   \\
         -120.293  &  5.593     &  91.499   \\
         -0.308    &  91.499    &  0.062    \\
    \end{array}
    \right) \nonumber \\
    \bm{\Sigma}_\mathbf{T} &= \left(
    \begin{array}{ccc}
         12.364 	&  -9.272  &  -13.966   \\
         -9.272 	&  0.728   &  2.215     \\
         -13.966    &  2.215   &  3.989     \\
    \end{array}
    \right)
    \label{eq:uncertainty_ratio}
\end{align}
The rotation uncertainty along the y-axis and the translation uncertainty along the x- and z-axis under the highway scenario are larger than the uncertainty under the urban environment.
Because more LiDAR and stereo points on the ground under the highway scenario are used so that the translation uncertainty along the y-axis and the rotation uncertainty along the x- and z-axis is smaller than the uncertainty under the urban environment.
So the uncertainty analysis can provide us automatically the reliability of the estimated extrinsic parameters in all 6 DoF.

\section{CONCLUSION AND FUTURE WORK}
\label{sec:CONCLUSION AND FUTURE WORK}
In this work, we present TEScalib, a targetless LiDAR and stereo camera extrinsic self-calibration approach especially for autonomous driving platforms.
During the LiDAR-stereo calibration, a novel geometric information-based data association and objective function using point position and surface normal information are applied to estimate the extrinsic parameters accurately and robustly.
The left and right camera calibration uses image key points and LiDAR depth to build a photometric objective function, which is then minimized to realize a highly accurate stereo camera extrinsic calibration.
During driving, the so called co-calibration processes these two approaches iteratively.
Besides, we propose an uncertainty analysis, which reflects the reliability of the estimated extrinsic parameters under different driving scenarios.
Our TEScalib makes it possible for autonomous driving platforms to extrinsic calibrate the LiDAR and stereo camera setup online robustly and accurately only using urban driving scenarios without any manufactured calibration targets from bad initializations.
Using our uncertainty analysis, the quality of the estimated extrinsic parameters can be judged for different moving dimensions separately.
Besides that, this automatic judgement also allows us to reject bad driving scenarios automatically to improve the calibration results.
For the future work we will address three points: 
realizing the LiDAR-stereo calibration and the left and right camera calibration in one optimization jointly;
designing an online monitoring system to detect the calibration error and to decide whether a new calibration process should be triggered;
analyzing of constraints for each single dimension.
\section*{Acknowledgment}
This research is accomplished within the project “UNICARagil”.
We acknowledge the financial support for the project by the Federal Ministry of Education and Research of Germany (BMBF). 
%
%We especially thank [list of partners] for their contribution to this publication.
%

\printbibliography

@inproceedings{mutual,
  title={{{Automatic Targetless Extrinsic Calibration of A 3d Lidar and Camera by Maximizing Mutual Information}}},
  author={Pandey, Gaurav and McBride, James and Savarese, Silvio and Eustice, Ryan},
  booktitle={Proceedings of the AAAI Conference on Artificial Intelligence},
  volume={26},
  number={1},
  year={2012}
}

@article{motion_based,
  title={{{Motion-based Calibration of Multimodal Sensor Extrinsics and Timing Offset Estimation}}},
  author={Taylor, Zachary and Nieto, Juan},
  journal={IEEE Transactions on Robotics},
  volume={32},
  number={5},
  pages={1215--1229},
  year={2016},
  publisher={IEEE}
}

@inproceedings{ishikawa2018lidar,
  title={{{Lidar and Camera Calibration Using Motions Estimated by Sensor Fusion Odometry}}},
  author={Ishikawa, Ryoichi and Oishi, Takeshi and Ikeuchi, Katsushi},
  booktitle={2018 IEEE/RSJ International Conference on Intelligent Robots and Systems (IROS)},
  pages={7342--7349},
  year={2018},
  organization={IEEE}
}

@article{kang2020automatic,
  title={{{Automatic Targetless Camera--LIDAR Calibration by Aligning Edge With Gaussian Mixture Model}}},
  author={Kang, Jaehyeon and Doh, Nakju L},
  journal={Journal of Field Robotics},
  volume={37},
  number={1},
  pages={158--179},
  year={2020},
  publisher={Wiley Online Library}
}

@inproceedings{irie2016target,
  title={{{Targetless Camera-Lidar Extrinsic Calibration Using a Bagged Dependence Estimator}}},
  author={Irie, Kiyoshi and Sugiyama, Masashi and Tomono, Masahiro},
  booktitle={2016 IEEE International Conference on Automation Science and Engineering (CASE)},
  pages={1340--1347},
  year={2016},
  organization={IEEE}
}

@article{jeong2019road,
  title={{{The Road Is Enough! Extrinsic Calibration of Non-Overlapping Stereo Camera and Lidar Using Road Information}}},
  author={Jeong, Jinyong and Cho, Younghun and Kim, Ayoung},
  journal={IEEE Robotics and Automation Letters},
  volume={4},
  number={3},
  pages={2831--2838},
  year={2019},
  publisher={IEEE}
}

@article{cui2017line,
  title={{{Line-based Registration of Panoramic Images and LiDAR Point Clouds for Mobile Mapping}}},
  author={Cui, Tingting and Ji, Shunping and Shan, Jie and Gong, Jianya and Liu, Kejian},
  journal={Sensors},
  volume={17},
  number={1},
  pages={70},
  year={2017},
  publisher={Multidisciplinary Digital Publishing Institute}
}

@article{shi2019extrinsic,
  title={{{Extrinsic Calibration And Odometry for Camera-LiDAR Systems}}},
  author={Shi, Chenghao and Huang, Kaihong and Yu, Qinghua and Xiao, Junhao and Lu, Huimin and Xie, Chenggang},
  journal={IEEE Access},
  volume={7},
  pages={120106--120116},
  year={2019},
  publisher={IEEE}
}

@book{bradski2008learning,
  title={{{Learning OpenCV: Computer Vision with the OpenCV Library}}},
  author={Bradski, Gary and Kaehler, Adrian},
  year={2008},
  publisher={" O'Reilly Media, Inc."}
}

@article{geiger2013vision,
  title={{{Vision Meets Robotics: The Kitti Dataset}}},
  author={Geiger, Andreas and Lenz, Philip and Stiller, Christoph and Urtasun, Raquel},
  journal={The International Journal of Robotics Research},
  volume={32},
  number={11},
  pages={1231--1237},
  year={2013},
  publisher={Sage Publications Sage UK: London, England}
}

@article{munoz2020targetless,
  title={{{Targetless Camera-LiDAR Calibration in Unstructured Environments}}},
  author={Mu{\~n}oz-Ba{\~n}{\'o}n, Miguel {\'A}ngel and Candelas, Francisco A and Torres, Fernando},
  journal={IEEE Access},
  volume={8},
  pages={143692--143705},
  year={2020},
  publisher={IEEE}
}

@INPROCEEDINGS{sgbm2005,
  author={H. {Hirschmuller}},
  booktitle={2005 IEEE Computer Society Conference on Computer Vision and Pattern Recognition (CVPR'05)}, 
  title={{{Accurate and Efficient Stereo Processing by Semi-Global Matching and Mutual Information}}}, 
  year={2005},
  volume={2},
  number={},
  pages={807-814 vol. 2},
  doi={10.1109/CVPR.2005.56},
  ISSN={1063-6919},
  month={June}
}

@inproceedings{xiao2017accurate,
  title={{{Accurate Extrinsic Calibration Between Monocular Camera and Sparse 3D Lidar Points without Markers}}},
  author={Xiao, Zhipeng and Li, Hongdong and Zhou, Dingfu and Dai, Yuchao and Dai, Bin},
  booktitle={2017 IEEE Intelligent Vehicles Symposium (IV)},
  pages={424--429},
  year={2017},
  organization={IEEE}
}

@article{wang2020soic,
  title={{{SOIC: Semantic Online Initialization and Calibration for LiDAR and Camera}}},
  author={Wang, Weimin and Nobuhara, Shohei and Nakamura, Ryosuke and Sakurada, Ken},
  journal={arXiv preprint arXiv:2003.04260},
  year={2020}
}

@article{taylor2016motion,
  title={{{Motion-based Calibration of Multimodal Sensor Extrinsics and Timing Offset Estimation}}},
  author={Taylor, Zachary and Nieto, Juan},
  journal={IEEE Transactions on Robotics},
  volume={32},
  number={5},
  pages={1215--1229},
  year={2016},
  publisher={IEEE}
}

@article{spartial_temperal,
  title={Spatiotemporal camera-LiDAR calibration: A targetless and structureless approach},
  author={Park, Chanoh and Moghadam, Peyman and Kim, Soohwan and Sridharan, Sridha and Fookes, Clinton},
  journal={IEEE Robotics and Automation Letters},
  volume={5},
  number={2},
  pages={1556--1563},
  year={2020},
  publisher={IEEE}
}

@inproceedings{wang2017stereo,
  title={Stereo DSO: Large-scale Direct Sparse Visual Odometry with Stereo Cameras},
  author={Wang, Rui and Schworer, Martin and Cremers, Daniel},
  booktitle={Proceedings of the IEEE International Conference on Computer Vision},
  pages={3903--3911},
  year={2017}
}

@ARTICLE{Spatiotemporal,
  author={Park, Chanoh and Moghadam, Peyman and Kim, Soohwan and Sridharan, Sridha and Fookes, Clinton},
  journal={IEEE Robotics and Automation Letters}, 
  title={Spatiotemporal Camera-LiDAR Calibration: A Targetless and Structureless Approach}, 
  year={2020},
  volume={5},
  number={2},
  pages={1556-1563},
  doi={10.1109/LRA.2020.2969164}
}

@ARTICLE{HandEyeCalib,
  author={Nguyen, Huy and Pham, Quang-Cuong},
  journal={IEEE Transactions on Robotics}, 
  title={On the Covariance of AX=XB}, 
  year={2018},
  volume={34},
  number={6},
  pages={1651-1658},
  doi={10.1109/TRO.2018.2861905}
}

@article{yuan2021pixel,
  title={Pixel-level Extrinsic Self Calibration of High Resolution LiDAR and Camera in Targetless Environments},
  author={Yuan, Chongjian and Liu, Xiyuan and Hong, Xiaoping and Zhang, Fu},
  journal={arXiv preprint arXiv:2103.01627},
  year={2021}
}

\end{document}